\documentclass[journal]{IEEEtran}
\usepackage{cite}
\usepackage{amsmath}

\usepackage{array}
\newcolumntype{P}[1]{>{\centering\arraybackslash}p{#1}}
\usepackage{mathtools}
\usepackage{stfloats} 
\usepackage{threeparttable}
\usepackage{booktabs} 
\usepackage{tabularx} 
\usepackage{lipsum} 
\usepackage{amsmath}
\usepackage{amssymb}
\usepackage[hidelinks,hypertexnames=false]{hyperref}
\usepackage{graphicx}

\usepackage{adjustbox}
\usepackage{stfloats}
\usepackage{float}
\usepackage{titlesec}
\usepackage{adjustbox}

\titlespacing{\section}{0pt}{*1.0}{*1}
\titlespacing{\subsection}{0pt}{*1.0}{*0.5}

\begin{document}
\title{Unified Pedestrian Path Prediction Using Inverse Reinforcement Learning}
\author{
\vspace{-10pt}
    \IEEEauthorblockN{Šimon Sukup\textsuperscript{1}, Ariyan Bighashdel\textsuperscript{1,2}, and Pavol Jancura\textsuperscript{1}} \\
    \textsuperscript{1}Eindhoven University of Technology, Eindhoven, The Netherlands \\
    \textsuperscript{2}Delft University of Technology, Delft, The Netherlands \\
    Corresponding author: \href{mailto:s.sukup@student.tue.nl}{s.sukup@student.tue.nl}
    \vspace{-7mm} 
  }
\IEEEaftertitletext{%
    \vspace{-0.8em}
    \begin{center}
        \includegraphics[width=0.99\textwidth]{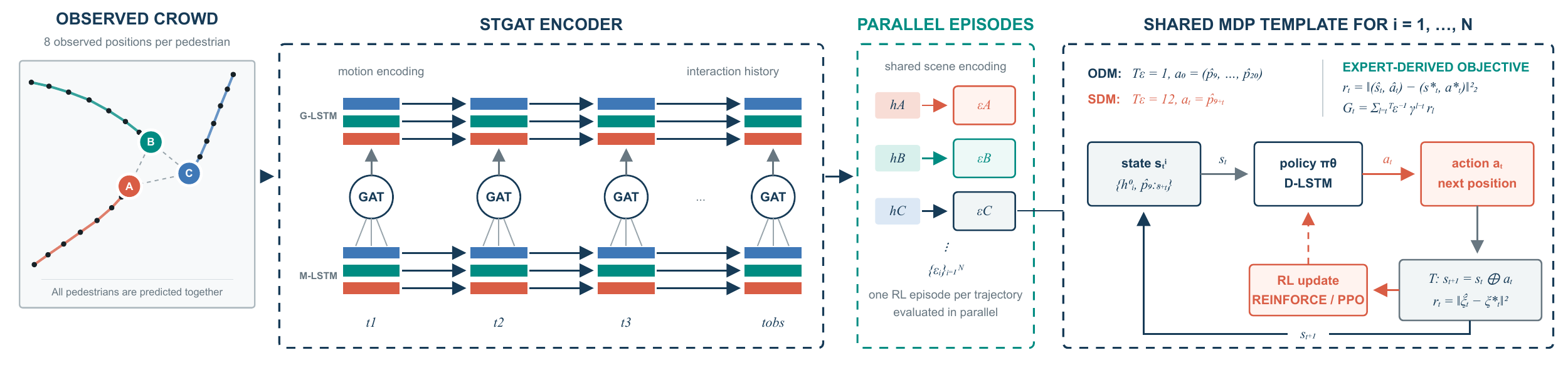}
        \refstepcounter{figure}\label{fig:trajectory_overview}
        \vspace{-0.45em}
        \parbox{0.975\textwidth}{\footnotesize
            \textbf{Fig.~\thefigure.} Proposed RL formulation. STGAT~\cite{stgat} jointly encodes and predicts all pedestrians in a scene. For optimization, the batched output is decomposed into parallel per-pedestrian episodes $\{\varepsilon_i\}_{i=1}^{N}$ governed by one shared MDP template. Each state combines the encoded scene context with previously predicted positions for pedestrian $i$. The D-LSTM emits either one trajectory-level action (ODM) or 12 next-position actions (SDM); expert demonstrations define the distance-based return used by REINFORCE or PPO.}
    \end{center}
    \vspace{0.35em}
}
\maketitle
\begin{abstract}

Pedestrian path prediction (PPP) is crucial for enhancing the safety of autonomous vehicles (AVs) and advanced driving assistance systems (ADAS). Previous studies explored various learning task formulations for PPP and performed comparative analyses of the formulations on shallow neural networks. However, such studies did not extend to the more complex deep learning models.
This paper utilizes a unified framework for PPP and adapts a widely used baseline model, the Spatial-Temporal Graph Attention Network (STGAT), to different learning task formulations. New state and action definitions specific to STGAT are introduced. 
The newly designed learning task formulations achieve performance improvement on all chosen benchmark datasets compared to the standard supervised learning task. 
This paper demonstrates, for the first time, that using alternative formulations, as opposed to the widely used supervised learning, can improve an advanced PPP model under a unified PPP framework. The designed formulations have a significant potential for improving a wide range of state-of-the-art (SOTA) architectures.


\end{abstract}
\noindent\textbf{Code:} \url{https://github.com/Simsuk/IRL_STGAT}\par
\section{Introduction}

\IEEEPARstart{H}{uman} errors contribute significantly to road traffic accidents \cite{error}. Advancements in AVs and ADAS have increased the potential to reduce such human-caused incidents and enhance road safety. A critical component of realizing this potential is the ability to accurately predict pedestrian behaviour, which allows AVs to comprehend their environment more effectively. The PPP challenge lies in the crowd interactions and the underlying decision-making process of each pedestrian. These aspects necessitate various learning task formulations introduced throughout the literature \cite{survey}.

To address the problem of objective comparison of different learning task formulations,  Lemmens et al. introduced unified framework for PPP \cite{fram}. The framework defines PPP as a decision-making problem in the form of a Markov decision process (MDP). Lemmens et al. demonstrated improvement in the performance of a shallow neural network policy when transferring from standard supervised learning settings to inverse reinforcement learning (IRL) settings or more advanced reinforcement learning (RL) algorithms \cite{fram}. However, the work has not extended the framework to more advanced deep learning architectures that are predominant in the field of PPP. 


This paper addresses an unresolved research question of whether the framework can enhance more advanced models. It contributes by extensive modification of the framework to adopt the STGAT \cite{stgat}, a widely used advanced architecture for PPP. STGAT was utilized as a stochastic and deterministic policy through the unified PPP framework while using advanced RL algorithms. The framework was extended with novel state and action definitions to design new learning task formulations specific to STGAT, including supervised learning for sequential decision-making and modified stochastic policy gradient one-time decision-making. Furthermore, this paper introduced two value network architectures specific to STGAT based on the designed state-actions pairs. The last contribution includes an ablation study on a newly defined sequential decision-making process.
This paper demonstrates that transitioning from supervised learning tasks to inverse reinforcement learning (IRL) with advanced algorithms can improve the performance of more complex PPP models. The current results and further extension of the framework can potentially lead to the development of better PPP models based on the graph attention network (GAT) architecture and improve a wide range of current SOTA architectures for PPP.

\section{Related Work}
The STGAT leverages deep learning techniques for time series forecasting and interaction modelling \cite{stgat}. This sections, therefore, covers the preceding works in the field that led to the development of STGAT and compares it to other literature. Furthermore, the motivation for the use of the unified pedestrian path prediction framework is explained with respect to previous works in the field.
\subsection{Advances in Pedestrian Path Prediction and Interaction Modelling}
Early PPP models lacked in generalization ability and scalability\cite{PhysRevE.51.4282}, \cite{eth} leading to the rise of deep learning methodologies. Significant attention in PPP has been given to the use of recurrent neural networks (RNNs), long-short-term-memory networks (LSTMs) and more advanced architectures based on gated recurrent units (GRUs) or transformers \cite{8794474},\cite{9008118},\cite{8626436},\cite{Liang_2019_CVPR}, \cite{Yuan2021AgentFormerAT}. 
Key studies, such as those by Alahi et al., have utilized LSTMs to address temporal dependencies within the data \cite{7780479}.

Aside from temporal dependencies, a series of studies have shown that understanding social interactions play a crucial role in PPP. The development of methods for modelling intentions and social behaviour drove research into techniques such as spatio-temporal graphs which integrate both spatial and temporal data. Key contributions to this branch of the field include the Social Attention and Trajectron models which use LSTMs to manage structured sequence data effectively \cite{Vemula2017SocialAM},\cite{trajectron}.

Recent publications have shown the importance of graph-structured networks in modelling interactions, like Graph Convolutional Networks (GCNs) \cite{9774877} and GATs \cite{velickovic2018graph}.
STGAT, or Spatio-Temporal Graph Attention Network, represents a significant improvement while building on the GAT methodology for modelling of dynamic interactions among pedestrians \cite{stgat}. This model is distinct in its ability to not only account for spatial interactions among individuals but also to capture the temporal evolution of these interactions over time. At the core of STGAT is the integration of GAT with LSTM networks. The GAT component effectively aggregates the hidden states from multiple LSTMs, allowing the model to assign variable importance to different pedestrians based on their proximity and relative significance in the scene. 
STGAT is utilized in this paper due to its wide use in the field and high performance, as well as being the foundation for current SOTA architectures.

\subsection{Unified Pedestrian Path Prediction Framework}
While the majority of state-of-the-art methods for PPP featured learning directly from ground truth data through supervised learning, there have been instances of research that utilized alternative formulations.

Such a significant line of research in this domain involves generative models. Gupta et al. introduced a generative adversarial network (GAN) formulation which is based on a min-max game between the trajectory generator and a discriminator where the generator distinguishes between real and generated trajectories \cite{DBLP:journals/corr/abs-1803-10892}. Extensions of this include the use of IRL while utilizing algorithms based on an algorithm based on generative adversarial imitation learning algorithm (GAIL) \cite{GAIL}, such as  SA-GAIL \cite{sa_gail}. Other works used generative adversarial imitation learning that involves learning from expert demonstrations through rewards received from interactions with the environment.  


While these distinct models with different learning formulations are often presented with improvement, it is often not clear if the improvement is made by the model architecture and modules or the new learning task formulation. To address this challenge,  Lemmens et al. introduced a unified PPP framework \cite{fram}. The original paper \cite{fram} utilized a simple shallow neural network which did not guarantee that the found performance improvement with advanced RL algorithms would be present while using more advanced models. Our work therefore extends the framework by designing new state-action formulations and adopting widely used advanced deep learning architecture STGAT \cite{stgat}. 



\section{Methods}

The extension of the unified framework required extensive modification of the codebase while introducing novel state-action definitions, value network architecture and loss definitions in order to utilize the architecture of the STGAT.  While the STGAT paper \cite{stgat} utilized simple supervised learning, this work treats path prediction as a reinforcement learning task where different framework settings lead to one of the newly designed formulations.

\subsection{Reinforcement Learning Problem Setup}

RL is a machine learning paradigm where an agent learns to make decisions by interacting with an environment, modelled as a Markov decision process (MDP). In RL, the agent's goal is to discover a policy, a strategy for choosing actions based on states, that maximizes the cumulative reward over time. 
MDP is defined by the following components. \(\mathcal{S}\) represents the state space, \(\mathcal{A}\) is the action space, the reward function \( \mathcal{R}: \mathcal{S} \times \mathcal{A} \rightarrow \mathbb{R}\) assigns a reward value \( \mathcal{R}(s, a) \) for taking action \( a \) in state \( s \), the state transition function \(\mathcal{T}: \mathcal{S} \times \mathcal{A} \rightarrow \mathcal{S}\) describes how the state changes, after taking an action \( a \) in the state \( s \). The discount factor \(\gamma\) is a value between 0 and 1 that determines the weight of future rewards compared to immediate ones. Finally, the initial state distribution \( b_0 : \mathcal{S} \rightarrow [0,1] \) gives the probability \( b_0(s) \) of starting in state \( s \).


The decision of an agent can either be based on stochastic policy $\pi_\theta\left(a \mid s\right):S \times A \rightarrow[0,1]$, which assigns probability to a state-action pair,  or deterministic policy $\mu_\theta\left(a \mid s\right): S \rightarrow A$, which assigns single action to every state. In this work, the policy is a mapping parameterized by the chosen learnable parameters of STGAT, denoted $\theta$. 

In this paper, the goal is to observe the trajectory at the initial time steps 
\(1\) to \(T_o^\tau=8\) of all pedestrians in the observed scene and forecast the future trajectory steps steps at times $t^\tau$ for \(T_o^\tau+1 \leq t^\tau\leq T_o^\tau+T_f^\tau\), where \(T_f^\tau=12\). For each $t^\tau$ the position of the pedestrian is represented by a vector of coordinates in the scene \({p}_{t^\tau}\) =\((x_{t^\tau}, y_{t^\tau})\). The observed trajectory is then denoted as \(\tau_f = \{p_{1}, \ldots, p_{T_o^\tau}\}\) while the forecasted trajectory \(\hat{\tau}_f = \{\hat{p}_{T_o^\tau+1}, \ldots, \hat{p}_{T_o^\tau+T_f^\tau}\}\), where \(\hat{p}_{t^\tau}\) denotes a predicted position at time step  \( T_0^\tau+1 \leq t^\tau \leq T_f^\tau\).

Let one-time decision-making (ODM) be a process where the initial state of observed steps is followed by a single action that represents all $12$ predicted steps for the pedestrian. The agent receives a single reward from the environment. Let $\epsilon$ denote the set of states, actions and rewards for a trajectory, also called an episode. For ODM, there is single step in episode  \(T^\varepsilon = 1 \) and $\varepsilon=\left\{\left(s_0, a_0, r_0\right),\left(s_1\right)\right\}$. Using ODM, the single action incorporates all pedestrian steps from \(T_o^\tau+1\) to \( T_f^\tau\) and action becomes \(a_0=\left(\hat{x}_{T_o^\tau+1},\hat{y}_{T_o^\tau+1}, \hat{x}_{T_o^\tau+2},\hat{y}_{T_o^\tau+2},\ldots, \hat{x}_{T_o^\tau+T_f^\tau}, \hat{y}_{T_o^\tau+T_f^\tau}\right)\) where $\hat{x}_{t^\tau}, \hat{y}_{t^\tau}$ are the predicted coordinates by the policy.

While ODM is standardly used in the literature, it does not appropriately reflect the decision-making process of a pedestrian. The SDM is therefore employed, which considers the pedestrian steps as separate actions and each episode then becomes $\varepsilon=\left\{\left(s_0, a_0\right), \ldots,\left(s_{T^{\varepsilon}-1}, a_{T^{\varepsilon}-1}\right),\left(s_{T^{\varepsilon}}\right)\right\}$ with \(T^\varepsilon = 12 \) steps in the episode. The action becomes \(a_{t^\epsilon}=\left(\hat{x}_{T_o^\tau+1 +t^{\epsilon}},\hat{y}_{T_o^\tau+1+t^{\epsilon}}\right)  \) for any step in the episode $0 \leq t^\epsilon<T^\epsilon$.
\\
Compared to RL, in IRL the real rewards are unknown and are estimated from the expert behaviour. In PPP, the expert behaviour is observed in the ground truth trajectories. Taking the IRL approach, let the estimated reward function be defined by a distance function, squared L2 norm,
\begin{equation}
\mathcal{R}(s_{t^\epsilon}, a_{t^\epsilon})=
r_{t^\epsilon}=\|(\left(\hat{s}_{t^\epsilon}, \hat{a}_{t^\epsilon}\right),\left(s_{t^\epsilon}, a_{t^\epsilon} \right)\|_2^2, 
\label{eq:reward}
\end{equation}
where index $t^\epsilon$ denotes a time step in the episode, $\hat{s}_{t^\epsilon}, \hat{a}_{t^\epsilon}$ are the current state and action coming from the predicted trajectory according to policy. The $s_{t^\epsilon}, a_{t^\epsilon}$ denote ground truth states and actions. Commonly, the rewards received when starting from a specific state are discounted in order to allow a choice of focus on future states. This results in the definition of discounted return
\begin{equation}
G\left(\varepsilon \mid s_{t^\epsilon}\right)=\sum_{l=t^\epsilon}^{T^\epsilon-1} \gamma^{l-t^\epsilon} r_l.
\label{eq:discounted}
\end{equation}
The state value function is defined as,
\begin{equation}
V(s ; \theta)=\mathbb{E}_{\left(\varepsilon \mid s_{t^\epsilon}=s\right) \sim p_\theta}\left[G\left(\varepsilon \mid s_{t^\epsilon}=s\right)\right]
\label{eq:value_f}
\end{equation}
The $p_\theta$ denotes either stochastic policy $\pi_\theta $  or deterministic policy $\mu_\theta$ depending on the experiment. $\mathbb{E}_{\left(\varepsilon \mid s_{t^\epsilon}=s\right)\sim p_\theta}$ denotes expectation over episodes starting from state $s$, when following the policy $p_\theta$. According to the policy gradient theorem\cite{reinf}, the parameters are updated according to the objective function, which is the expected value of initial states
\begin{equation}
J(\theta)=\mathbb{E}_{s_0 \sim b_0(s)} V(s_0 ; \theta).
\end{equation}
where $\mathbb{E}_{s \sim b_0(s)}$ denotes expectation over initial states sampled from the initial state distribution. The parameters of the policy are then updated using a standard stochastic gradient descent algorithm with learning rate $\alpha$,

\begin{equation}
\theta_{k+1}=\theta_k+\alpha \nabla_\theta J\left(\pi_{\theta_k}\right).
\end{equation}

\subsection{STGAT Architecture and Implementation}
\label{sec:stgat_implementation}
STGAT comprises several modules: the encoder, intermediate state vectors, and the decoder. The encoder uses two separate LSTM modules, one for modelling temporal correlations between interactions in time, denoted as G-LSTM, and the second M-LSTM to model the spatial interactions. 
The STGAT models the interactions in the encoder using only one scene of pedestrians at a time. 
Noise is added to the outputs of the encoder for multimodality.
\\
The decoder comprises of a single LSTM (D-LSTM) that takes the intermediate state vectors as input sequences and outputs the next steps of each pedestrian separately.  
 In the first two phases of training, taking 250 epochs, the STGAT learns to reconstruct the observed trajectory through the encoder. In the third phase, the model trains the whole pipeline, including the D-LSTM in 150 epochs.
After pretraining on the first 2 phases, the model checkpoint was reused for the third phase, where the formulations described in Section~\ref{sec:ODM} and Section~\ref{sec:SDM} were used.
\subsection{ODM Formulations}
\label{sec:ODM}
\subsubsection{SL-MSE}
The supervised learning mean-square-error (SL-MSE) formulation corresponds to the ODM process. Because STGAT considers the whole scenes of pedestrians as single input in order to model the interactions, the states are defined as the set of 8 observed steps of all pedestrians in the scene. The action is defined as the next 12 steps of a single pedestrian. After the first action, the final state becomes the whole trajectory of 20 steps. For SL-MSE the discount factor is $\lambda=0$. To enable multimodality, Gaussian noise was incorporated into the input states for training and evaluation.
Using equations (\ref{eq:reward}), (\ref{eq:discounted}) and  (\ref{eq:value_f}) the objective function gradient simplifies to
\begin{align}
\nabla_{\theta} J(\theta)
&= \mathbb{E}_{s_0 \sim b_0(s_0),\, (\varepsilon \mid s_0) \sim \pi_\theta}
\Bigg[\smashoperator{\sum_{t^\tau=T_o^\tau+1}^{T_o^\tau+T_f^\tau}}
\nabla_{\theta} \Big((\hat{x}_{t^\tau}-x_{t^\tau})^2 \notag \\
&\qquad + (\hat{y}_{t^\tau}-y_{t^\tau})^2\Big)\Bigg],
\end{align}
where $t^\tau$ denotes the corresponding timestamp in the trajectory, $\hat{x_{t^\tau}}, \hat{y_{t^\tau}}$ are predicted coordinates, and $x_{t^\tau}, y_{t^\tau}$. 
\subsubsection{SL-MSE-SPG}
The stochastic gradient policy variant of the supervised learning formulation required modification of the model. Because stochastic policies in RL commonly use Gaussian distributions to sample every next step of each pedestrian sequentially, the noise added to the outputs of the encoder of STGAT was fixed for each initialization of the model. This allowed finding the true probabilities by sampling from output Gaussian distributions with a fixed standard deviation of 0.05. This was necessary in order to preserve the model capacity, which is dependent on the encoded state's dimensions. The states and actions were defined in the same way as in the case of SL-MSE. Using the  stochastic policy gradient theorem\cite{reinf}, (\ref{eq:reward}), (\ref{eq:discounted}) and  (\ref{eq:value_f}) results in an expression
\begin{align}
&\nabla_{\theta} J(\theta) =\mathbb{E}_{s_0 \sim b_0(s_0),\, (\varepsilon \mid s_0) \sim \pi_\theta} \mkern-1mu \Bigg[ \sum_{t=T_o^\tau +1}^{T_o^\tau+ T_f^\tau} \mkern-11mu\Big((\hat{x}_{t^\tau} - x_{t^\tau})^2 \notag \\
& + (\hat{y}_{t^\tau} - y_{t^\tau})^2 \Big)
 \times \nabla_{\theta} \left( \sum_{t^\tau=T_o^\tau +1}^{T_o^\tau+ T_f^\tau} \log \pi_{\theta} \left( \hat{a_{t^\tau}} \mid \hat{s_{t^\tau}} \right) \right) \Bigg],
\end{align}
where $ \log \pi_{\theta} \left( \hat{a_{t^\tau}} \mid \hat{s_{t^\tau}} \right) $ denotes a logarithm of conditional probability in trajectory time step $t^\tau$ when choosing an action given the current state.
\subsection{SDM Formulations (BEP Extension)}
\label{sec:SDM}
Scenes involving pedestrians inherently represent multi-agent scenarios where each pedestrian interacts with surrounding neighbours. The SDM (Single Decision Maker) approach simplifies the problem by focusing on individual pedestrian decision-making, treating each trajectory as an independent episode and keeping the interactions in the observed state.
In STGAT, the initial encoded state of the whole scene is used, but subsequent predictions rely only on the hidden state of the D-LSTM, ignoring interactions with other agents. Therefore, the SDM formulation reduces to a single-agent problem, where pedestrian trajectories are treated independently, and all observed steps in the scene serve as an initial state for each pedestrian.
\subsubsection{SL-MSE-SDM}
In order to utilize a step-wise decision-making process of the pedestrian, the action was defined as the single step of a pedestrian in the trajectory.
The next state was defined to be a vector consisting of the previous state and current action. The number of steps in an episode for SDM is $T^{\epsilon}=12$ and the episode time step is $0 \leq t^\epsilon<T^\epsilon$. To be mathematically sound, let action $a_{t^\epsilon}\in \mathbb{R}^{2}$  such that \(a_{t^\epsilon}=\left(\hat{x}_{T_o^\tau+1 +t^{\epsilon}},\hat{y}_{T_o^\tau+1+t^{\epsilon}}\right)  \). 
Let the initial state $s_0 \in \mathbb{R}^{2N_0}$, where $N_0$ is a number of x-y pairs of observed trajectories of all pedestrians in the scene, each having 8 steps. Let $s_{t^\epsilon} \in \mathbb{R}^{2N_{t^\epsilon}}$ be state vector $s_{t^\epsilon}=(x_{s,0}, y_{s,0}, ..., x_{s,N_{t^\epsilon}}, y_{s,N_{t^\epsilon}})$ comprising of $2N_{t^\epsilon}$ components. The number of x-y pairs $N_{t^\epsilon}$ of state $s_{t^\epsilon}$ at the time step of the episode $t^\epsilon$  is therefore dependent on the number of pedestrians in the observed scene.  The transition function can then be defined as mapping $\mathcal{T} : \mathbb{R}^{N} \times \mathbb{R}^2 \to \mathbb{R}^{N+2}$ such that
\begin{align}
s_{t^\epsilon+1}&=\mathcal{T}(s_{t^\epsilon}, a_{t^\epsilon}) = (x_{s,0}, y_{s,0}, ..., \notag \\
& x_{s,N_{t^\epsilon}}, y_{s,N_{t^\epsilon}},  \hat{x}_{T_o^\tau+1 +t^{\epsilon}},\hat{y}_{T_o^\tau+1+t^{\epsilon}}).
\end{align}
The termination state is reached when \(t^\epsilon=T^\epsilon-1\). 
Taking into account the definition of the reward function, the resulting policy gradient becomes
\begin{align}
\label{eq:SDM}
& \nabla_ {\theta} J(\theta)  =\mathbb{E}_{s_0 \sim b_0\left(s_0\right),\left(\varepsilon \mid s_0\right) \sim \pi_\theta} \notag \\
&\Bigg[ \nabla_{\theta}\sum_{t^\tau=T_o^\tau +1}^{T_o^\tau+ T_f^\tau}  \gamma^{t^\tau-T_o^\tau -1}  \mkern-14mu
\sum_{l'=T_o^\tau+1 }^{t^\tau} \mkern-14mu
 \left((\hat{x_{l'}} - x_{l'})^2 + (\hat{y_{l'}} - y_{l'})^2\right) \Bigg].
\end{align}
\section{Experimental Settings}

\subsection{Datasets}
ETH \cite{eth} and UCY \cite{ucy} are standard benchmarking datasets for PPP also used in this work. The datasets include 2D coordinates and time steps of each step. Pedestrians are divided into scenes, which are sets of observed and next-step trajectories in a single location. Such scenes therefore contain only pedestrians who were present at the location at the same time and possibly interacted. The ETH dataset consists of two subsets, ETH and Hotel.  Similarly, the UCY dataset consists of three subsets, Zara1, Zara2 and Univ. The results are reported as per individual subsets of the two datasets. The subsets are further divided into training, validation and testing sets. The validation set was generated using a leave-one-out strategy where four-fifths of the original training set of trajectories were employed as the train set while the remaining fifth of trajectories was used for evaluation.

\subsection{Evaluation and Metrics}
The Average Displacement Error (ADE) and Final Displacement Error (FDE) were used as the metrics most widely used in the field of PPP. The mean of output distribution STGAT was used as an action for ADE and FDE calculation to make SL-MSE and other formulations comparable.
Additionally, the prediction was made 20 times, and the ADE and FDE were calculated for each pedestrian. The minimum out of 20 runs per pedestrian averaged over all pedestrians in the set were reported as mean minFDE and mean minADE.
The results are reported as averages over 5 runs rounded to 2 decimals. Lower values indicate better prediction performance for all metrics in this paper.
\subsection{Training and Optimization}
In the first two pretraining phases, the STGAT model was trained five times with different initializations on each of the subsets of ETH and UCY. Similarly, every formulation and algorithm used for SL-MSE-SPG has been trained 5 times in the third learning phase. For each run of the training, a pre-trained model with unique initialization has been used.

The models have been fine-tuned for each algorithm and formulation on the ETH training and validation set.
In this paper, the optimal learning rate was kept identical over all modules of the STGAT in the third phase of learning. RL algorithms used included REINFORCE \cite{reinf} and proximal policy optimization (PPO) \cite{PPO}.

\subsection{Value function}
\label{sec:val}

REINFORCE with baseline and PPO utilized a value function approximation, which relied on a network predicting expected returns based on state inputs. To handle the large input space of full pedestrian scenes, this paper introduces a network comprising of a shallow 3-layer dense neural network paired with the STGAT's shared encoder, referred to as the full state baseline in the tables. This structure avoided extensive hyperparameter tuning and sped up processing by only updating the shallow layers and reusing the STGAT encoder.
Additionally, a value network that takes simplified subsets of the states was tested. This value network used a simplified input state of only a single pedestrian observed steps but maintained the same neural architecture as the full state network. This network is referred to as a simplified baseline.

\section{Results and Discussion}

\begin{table}[t] 
\centering
\setlength{\tabcolsep}{3pt}
\caption{Comparison of stochastic policy and deterministic policy results.}
\label{tab:model_predictions_stats}
\begin{adjustbox}{max width=\columnwidth}
\begin{tabular}{@{}l *{8}{c}@{}}
    \toprule
    & \multicolumn{4}{c}{SL-MSE} & \multicolumn{4}{c}{SL-MSE-SPG} \\
    \cmidrule(lr){2-5} \cmidrule(l){6-9}
    Formulation & ADE & minADE & FDE & minFDE & ADE & minADE & FDE & minFDE \\
    \midrule
    ETH & \textbf{0.95} & 0.89 & \textbf{1.84} & \textbf{1.69} & \textbf{0.95} & \textbf{0.87} & 1.85 & \textbf{1.69} \\
    Hotel & 0.61 & 0.55 & \textbf{1.19} & 1.07 & \textbf{0.60} & \textbf{0.52} & \textbf{1.19} & \textbf{1.06} \\
    Zara1 & \textbf{0.43} & \textbf{0.39} & \textbf{0.95} & \textbf{0.85} & 0.45 & 0.40 & 0.97 & \textbf{0.85} \\
    Zara2 & \textbf{0.37} & \textbf{0.32} & 0.79 & \textbf{0.70} & \textbf{0.37} & 0.35 & \textbf{0.77} & \textbf{0.70} \\
    Univ & \textbf{0.53} & \textbf{0.51} & \textbf{1.15} & \textbf{1.11} & 0.63 & 0.63 & 1.30 &  1.27 \\
    \midrule
    Average & \textbf{0.5801} & \textbf{0.5334} &\textbf{ 1.1824} & \textbf{1.0843} & 0.5987 & 0.5598 & 1.2197 & 1.1144 \\
    \bottomrule
\end{tabular}
\end{adjustbox}
\end{table}

Initially, this section examines both stochastic and deterministic policies in the context of ODM. The stochastic policy trained through REINFORCE, REINFORCE with baseline, PPO and PPO with and without baseline is evaluated. Subsequently, REINFORCE with baseline and PPO with baseline are compared while using different architectures of value function. Lastly, the SL-MSE-SDM is analyzed to assess SDM benefits on agent decision-making with STGAT.
\subsection{Comparing Deterministic and Stochastic ODM Policy with REINFORCE}
As can be seen in the Table \ref{tab:model_predictions_stats},  SL-MSE-SPG formulation has better or similar performance for the ETH, Hotel and Zara2. On the other hand, SL-MSE-SPG underperformed in the case of Zara1 and Univ subsets. This difference in performance can be attributed to the simplicity of the REINFORCE algorithm which was used in the case of SL-MSE-SPG. As the simplest Monte Carlo stochastic policy gradient method, REINFORCE conveys high variance in gradient steps that lead to slower convergence and fluctuations in performance on the validation set, which was encountered during training. During the training, the best-performing iteration on the validation set often did not correspond to the best-performing iteration on the test set which was caused by the high variance in the loss.

This major drawback prevented the stochastic policy from reaching the same results as the deterministic policy on the fixed 150 epochs of the third phase of training resulting in all overall averages of SL-MSE-SPG higher. To assess the performance difference, further tuning of both algorithms on parameters such as batch size would be necessary.

\begin{table*}[htbp]
 \vspace{-4pt}
\vspace{1\intextsep}
\caption{Comparison of different Algorithm results with stochastic and deterministic policies.}
\centering
\begin{adjustbox}{width=1\linewidth}
\setlength{\tabcolsep}{3pt}
\begin{tabular}{l|c c c c|c c c c|c c c c|c c c c|c c c c}
    \toprule
    \multicolumn{1}{c}{} &
    \multicolumn{4}{c}{SL-MSE} &
    \multicolumn{4}{c}{SL-MSE-SPG} &
    \multicolumn{4}{c}{REINFORCE with Full State Baseline} &
    \multicolumn{4}{c}{PPO with  Full State Baseline} &
    \multicolumn{4}{c}{PPO without Baseline} \\
    \cmidrule(r){2-5}
    \cmidrule(r){6-9}
    \cmidrule(r){10-13}
    \cmidrule(r){14-17}
    \cmidrule(r){18-21}
    \textbf{Dataset} & ADE & minADE & FDE & minFDE & ADE & minADE & FDE & minFDE & ADE & minADE & FDE & minFDE & ADE & minADE & FDE & minFDE & ADE & minADE & FDE & minFDE \\
    \toprule
    ETH   & 0.95 & 0.89 & 1.84 & 1.69 & 0.95 & 0.87 & 1.85 & 1.69 & 0.93 & 0.86 & 1.82 & 1.67 & \textbf{0.91} & \textbf{0.84} & \textbf{1.79} & \textbf{1.64} & 0.94 & 0.86 & 1.85 & 1.69 \\
    Hotel & 0.61 & 0.55 & 1.19 & 1.07 & 0.60 & 0.52 & 1.19 & 1.06 & 0.61 & 0.57 & 1.21 & 1.11 & \textbf{0.54} & \textbf{0.51} & 1.11 & \textbf{1.03} & 0.56 & 0.52 & \textbf{1.10} & \textbf{1.03} \\
    Zara1 & 0.43 & 0.39 & 0.95 & 0.85 & 0.45 & 0.40 & 0.97 & 0.85 & \textbf{0.42} & \textbf{0.38} & 0.93 & 0.81 & \textbf{0.42} & \textbf{0.38} & \textbf{0.92} & \textbf{0.80} & 0.43 & 0.39 & 0.94 & 0.82 \\
    Zara2 & 0.37 & \textbf{0.32} & 0.79 & 0.70 & 0.37 & 0.35 & 0.77 & 0.70 & 0.34 & 0.34 & \textbf{0.71} & \textbf{0.68} & \textbf{0.33} & 0.34 & \textbf{0.71} & 0.69 & 0.34 & 0.35 & 0.75 & 0.71 \\
    Univ  & 0.53 & \textbf{0.51} & 1.15 & 1.11 & 0.63 & 0.63 & 1.30 & 1.27 & 0.53 & 0.54 & 1.15 & 1.13 & \textbf{0.52} & 0.53 & \textbf{1.13} & \textbf{1.11} & 0.56 & 0.56 & 1.19 & 1.17 \\
    \toprule
    Average & 0.5801 & 0.5334 & 1.1824 & 1.0843 & 0.5987 & 0.5598 & 1.2197 & 1.1144 & 0.563 &  0.535 & 1.168 & 1.084 & \textbf{0.5460} & \textbf{0.5227} & \textbf{1.1335} & \textbf{1.0551} & 0.5644 & 0.5370 & 1.1675 & 1.0844 \\
    \bottomrule
    
\end{tabular}
\end{adjustbox}
\label{tab:advanced_alg}
 \vspace{-12pt}
\end{table*}

\begin{table*}[htbp]
 \vspace{-2pt}
\vspace{1\intextsep}
\caption{Comparison of different value network implementations.}
\centering
\begin{adjustbox}{width=1\linewidth}
\setlength{\tabcolsep}{3pt}
\begin{tabular}{l|c c c c|c c c c|c c c c|c c c c}
    \toprule
    \multicolumn{1}{c}{} &
    \multicolumn{4}{c}{REINFORCE with Simplified Baseline} &
    \multicolumn{4}{c}{REINFORCE with Full State Baseline} &
    \multicolumn{4}{c}{PPO with Simplified Baseline} &
    \multicolumn{4}{c}{PPO with Full State Baseline} \\
    \cmidrule(r){2-5}
    \cmidrule(r){6-9}
    \cmidrule(r){10-13}
    \cmidrule(r){14-17}
    \textbf{Dataset} & ADE & minADE & FDE & minFDE & ADE & minADE & FDE & minFDE & ADE & minADE & FDE & minFDE & ADE & minADE & FDE & minFDE \\
    \toprule
    ETH   & 0.92 & 0.85 & 1.81 & 1.66 & 0.93 & 0.86 & 1.82 & 1.67 & 0.92 & \textbf{0.84} & \textbf{1.79} & 1.63 & \textbf{0.91} & \textbf{0.84} & \textbf{1.79} & \textbf{1.64} \\
    Hotel & 0.67 & 0.63 & 1.35 & 1.26 & 0.61 & 0.57 & 1.21 & 1.11 &\textbf{ 0.53} & \textbf{0.50} & \textbf{1.10} & \textbf{1.03} & 0.54 & 0.51 & 1.11 & \textbf{1.03} \\
    Zara1 & 0.43  &  \textbf{0.38} & \textbf{0.92} & \textbf{0.80} & \textbf{0.42} & \textbf{0.38} & 0.93 & 0.81 & \textbf{0.42} & \textbf{0.38} & \textbf{0.92} & \textbf{0.80} & \textbf{0.42} & \textbf{0.38} & \textbf{0.92} & \textbf{0.80} \\
    Zara2 & 0.39 & 0.39 & 0.74 & 0.69 & 0.34 & \textbf{0.34} & \textbf{0.71} & \textbf{0.68} & \textbf{0.33} & 0.34 & 0.72 & 0.69 & \textbf{0.33} & \textbf{0.34} & \textbf{0.71} & 0.69 \\
    Univ  & 0.73 & 0.72 & 1.47 & 1.44 & 0.53 & 0.54 & 1.15 & 1.13 & \textbf{0.52} & \textbf{0.53} & \textbf{1.12} & \textbf{1.10 }& \textbf{0.52} & \textbf{0.53} & 1.13 & 1.11 \\
    \midrule
    Average & 0.6282&  0.5965 & 1.2603 & 1.1718 & 0.563 &  0.535& 1.168 & 1.084 & \textbf{0.5434} &  \textbf{0.5208}& \textbf{1.1304} & \textbf{1.0534} & 0.5460 & 0.5227 & 1.1335 & 1.0551 \\
    \bottomrule
\end{tabular}
\end{adjustbox}
\label{tab:value_func}
 \vspace{-10pt}
\end{table*}
\subsection{Comparing REINFORCE and PPO variations}
The hyperparameter tuning of parameters specific to each algorithm resulted in improvement on all datasets for PPO with full state baseline. This can be seen in Table \ref{tab:advanced_alg}. PPO with a full state baseline was the most advanced algorithm used in this paper. The highest performance gain for ADE was seen for ETH with 4\% decrease, and for Hotel with over 11\% decrease. A similar decrease was shown for FDE which ranged between 2\% and 10\% for all datasets. The second algorithm that resulted in the most prominent improvement in the metrics was the REINFORCE with a full state baseline. This training algorithm underperformed on several metrics compared to SL-MSE for Hotel and Univ while reaching the same or better results as PPO with a shared encoder baseline on 4 metrics for Zara1 and Zara2. This can be seen in Table \ref{tab:advanced_alg}. The reason why this algorithm underperformed on Univ and Hotel while improving performance on ETH compared to SL-MSE can be attributed to the fact that the model was fine-tuned on ETH. Further improvement on the Univ and Hotel could be achieved by fine-tuning on these data subsets separately.
Considering that mere use of the stochastic policy did not yield any overall improvement in Table \ref{tab:model_predictions_stats}, while PPO and REINFORCE with baseline did, this indicates the need for advanced RL algorithm to achieve improvement with regards to SL-MSE.

\subsection{Value Function Variations}
In parallel with the simulations with a value network with a shared STGAT encoder, a network taking simplified state input has been tested. The ablation study on the choice of the value network can be seen in Table \ref{tab:value_func}. Comparing the two REINFORCE implementations, the use of the full state through the value network with shared STGAT encoder resulted in an improvement of up to 27\%  ADE of Univ and between 8\% to 13\% in the case of metrics of Zara2 and Hotel. There has been a 1\% decline in the performance of metrics on ETH when using REINFORCE with full state baseline. The full state baseline therefore improved results compared to the simplified baseline for REINFORCE on average. A higher number of runs of the model training would be needed to justify the better performance of REINFORCE with a simplified baseline for FDE and minFDE metrics of Zara1.
In contrast, PPO with value network with the shared encoder did not result in major improvement or loss of performance for any of the metrics. The two value network approaches can therefore be considered equivalently well suited for the purpose of improving the model performance through the PPO in the unified framework. It needs to be noted that PPO with a simplified baseline takes subsets of states as input as has been explained in the Section~\ref{sec:val}.

Furthermore, when comparing Table~\ref{tab:advanced_alg}  and Table~\ref{tab:value_func} it becomes evident that PPO without baseline had worse performance compared to the shared encoder baseline and PPO with full state baseline. This indicates that using a separate network for modelling value function is a key towards improved performance.


\subsection{SDM Ablation Study (BEP extension)}
The results of SDM for deterministic policy in Table \ref{tab:SDM} 
showcase a decrease in metrics for all datasets compared to SL-MSE. Particularly for the $\lambda=0.4$, which the model was fine-tuned on, the best results were reached with decreasing between 2\% and 13\% and improvement on the majority of the rest of the metrics. When comparing the Table \ref{tab:advanced_alg}, the results for $\lambda=0.4$ outperform the PPO with a baseline on ETH while PPO performed better on the part of metrics of Zara2 and Univ in the range of 1\%-2\%.
Such improvement can be attributed to the choice of the SDM state-action formulation introduced in this work. Because of the cumulative nature of the state transitions, the loss function in equation (\ref{eq:SDM}) results in gradients that put more focus on learning the initial steps of the predicted trajectory. This indicates that the initial steps were decisive factors for good performance and focusing on the initial decision process of the pedestrian immediately after the observed 8 steps is key for model performance. This is further supported by the improvement of FDE considering that the final pedestrian location is highly affected by the initial decisions of the agent inference. Overall, by choosing an appropriate combination of the discount factor and the state-action definitions, an improvement can be achieved compared to the simple supervised method.

\begin{table*}[htbp]
 \vspace{-5pt}
\vspace{1\intextsep}
\caption{Comprehensive comparison of different algorithmic results with stochastic and deterministic policies.}
\centering
\begin{adjustbox}{width=1\linewidth}
\setlength{\tabcolsep}{3pt}
\begin{tabular}{l|c c c c|c c c c|c c c c|c c c c|c c c c}
    \toprule
    \multicolumn{1}{c}{} &
    \multicolumn{4}{c}{} &
    \multicolumn{16}{c}{SL-MSE-SDM} \\
    \cmidrule(r){6-21}
    \multicolumn{1}{c}{} &
    \multicolumn{4}{c}{SL-MSE-ODM} &
    \multicolumn{4}{c}{$\lambda = 0.2$} &
    \multicolumn{4}{c}{$\lambda = 0.4$} &
    \multicolumn{4}{c}{$\lambda = 0.6$} &
    \multicolumn{4}{c}{$\lambda = 0.8$} \\
    \cmidrule(r){2-5}
    \cmidrule(r){6-9}
    \cmidrule(r){10-13}
    \cmidrule(r){14-17}
    \cmidrule(r){18-21}
    Dataset & ADE & minADE & FDE & minFDE & ADE & minADE & FDE & minFDE & ADE & minADE & FDE & minFDE & ADE & minADE & FDE & minFDE & ADE & minADE & FDE & minFDE \\
    \toprule
    ETH   & 0.95 & 0.89 & \textbf{1.84} &\textbf{ 1.69 } & 0.91 & 0.84 & 1.9 & 1.75 & \textbf{0.90} &\textbf{ 0.85} & 1.88 & 1.74 & 0.91 & 0.86 & 1.87 & 1.73 & 0.93 & 0.87 & 1.87 & 1.73 \\
    Hotel & 0.61 & 0.55 & 1.19 & 1.07 & \textbf{0.52} & \textbf{0.48} & 1.12 & 1.02 & 0.53 & 0.49 & \textbf{1.09} & \textbf{1.01} & 0.53 & 0.49 & 1.1 & 1.03 & 0.55 & 0.51 & 1.12 & 1.03 \\
    Zara1 & 0.43 & 0.39 & \textbf{0.95} & \textbf{0.85} & 0.45 & 0.41 & 1.05 & 0.93 & \textbf{0.42} & \textbf{0.39} & \textbf{0.95} & 0.87 & 0.43 & 0.39 & \textbf{0.95} & 87 & 0.43 & 0.39 & 0.96 & 0.88 \\
    Zara2 & 0.37 & 0.32 & 0.79 & 0.70 & 0.35 & 0.32 & 0.81 & 0.71 & \textbf{0.34} & \textbf{0.31} & \textbf{0.76} & \textbf{0.69} & \textbf{0.34} & \textbf{0.31} & \textbf{0.76} & 0.70 & 0.35 & 0.32 & 0.77 & 0.70 \\
    Univ  & 0.53 & 0.51 & 1.15 & 1.11 & \textbf{0.52 }& \textbf{0.50} & 1.14 & \textbf{1.09} & \textbf{0.52} & \textbf{0.50} & \textbf{1.13} & \textbf{1.09} & \textbf{0.52} & 0.51 & \textbf{1.13} & 1.10 & \textbf{0.52} & 0.51 & 1.14 & 1.10 \\
    \toprule
    Average & 0.5801 & 0.5334 & 1.1824 & 1.0843 & 0.5511 & 0.5099 & 1.2030 & 1.1010 & \textbf{0.5439} & \textbf{0.5079} & \textbf{1.1646}& \textbf{1.0822} & 0.5476 & 0.5126 & 1.1650 & 1.0875 & 0.5545 & 0.5178 & 1.1704 & 1.0921 \\
    \bottomrule
\end{tabular}
\end{adjustbox}
\label{tab:SDM}
 \vspace{-10pt}
\end{table*}

\section{Conclusion}

In this work, STGAT was implemented in a unified PPP framework by designing new state action definitions, novel supervised SDM formulation and architectures for the value network. Algorithms including PPO, PPO without a baseline, REINFORCE and REINFORCE with a baseline were studied.

The use of ODM stochastic policy formulation with PPO with baseline and the designed supervised SDM formulation offer a substantial improvement in the performance of STGAT on all datasets. This result indicates that incorporating more sophisticated RL algorithms or learning task formulations that account for the stochastic nature of human movement offer an advantage in the learning process. This study therefore demonstrates that the unified framework can effectively enhance the performance of complex deep learning PPP architecture.

Further research on STGAT could investigate if additional improvement can be achieved through stochastic STGAT policy combined with SDM or by utilizing IRL and GAN formulations with a parameterized discriminator network that were not studied in this work. The key area of future work would be to apply the designed formulations and unified framework on a current SOTA architecture for PPP. 
Overall, the study shows a strong potential to improve a majority of the SOTA models in the PPP field through the unified PPP framework and advanced RL algorithms.
\bibliographystyle{IEEEtran}
\bibliography{main}

\begin{thebibliography}{10}
\providecommand{\url}[1]{#1}
\csname url@samestyle\endcsname
\providecommand{\newblock}{\relax}
\providecommand{\bibinfo}[2]{#2}
\providecommand{\BIBentrySTDinterwordspacing}{\spaceskip=0pt\relax}
\providecommand{\BIBentryALTinterwordstretchfactor}{4}
\providecommand{\BIBentryALTinterwordspacing}{\spaceskip=\fontdimen2\font plus
\BIBentryALTinterwordstretchfactor\fontdimen3\font minus
  \fontdimen4\font\relax}
\providecommand{\BIBforeignlanguage}[2]{{%
\expandafter\ifx\csname l@#1\endcsname\relax
\typeout{** WARNING: IEEEtran.bst: No hyphenation pattern has been}%
\typeout{** loaded for the language `#1'. Using the pattern for}%
\typeout{** the default language instead.}%
\else
\language=\csname l@#1\endcsname
\fi
#2}}
\providecommand{\BIBdecl}{\relax}
\BIBdecl

\bibitem{stgat}
Y.~Huang, H.~Bi, Z.~Li, T.~Mao, and Z.~Wang, ``{STGAT}: Modeling
  spatial-temporal interactions for human trajectory prediction,'' in
  \emph{Proceedings of the IEEE/CVF International Conference on Computer Vision
  (ICCV)}, 2019, pp. 6272--6281.

\bibitem{error}
D.~J. Fagnant and K.~Kockelman, ``Preparing a nation for autonomous vehicles:
  Opportunities, barriers and policy recommendations,'' \emph{Transportation
  Research Part A: Policy and Practice}, vol.~77, pp. 167--181, 2015.

\bibitem{survey}
N.~Sharma, C.~Dhiman, and S.~Indu, ``Pedestrian intention prediction for
  autonomous vehicles: A comprehensive survey,'' \emph{Neurocomputing}, vol.
  508, pp. 120--152, 2022.

\bibitem{fram}
J.~L.~A. Lemmens, A.~Bighashdel, P.~Jancura, and G.~Dubbelman, ``Unified
  pedestrian path prediction framework: A comparison study,'' in \emph{2023
  IEEE Intelligent Vehicles Symposium (IV)}, 2023, pp. 1--8.

\bibitem{PhysRevE.51.4282}
D.~Helbing and P.~Moln{\'a}r, ``Social force model for pedestrian dynamics,''
  \emph{Physical Review E}, vol.~51, pp. 4282--4286, 1995.

\bibitem{eth}
S.~Pellegrini, A.~Ess, K.~Schindler, and L.~Van~Gool, ``You'll never walk
  alone: Modeling social behavior for multi-target tracking,'' in \emph{2009
  IEEE 12th International Conference on Computer Vision}, 2009, pp. 261--268.

\bibitem{8794474}
Y.~Yao, M.~Xu, C.~Choi, D.~J. Crandall, E.~M. Atkins, and B.~Dariush,
  ``Egocentric vision-based future vehicle localization for intelligent driving
  assistance systems,'' in \emph{2019 International Conference on Robotics and
  Automation (ICRA)}, 2019, pp. 9711--9717.

\bibitem{9008118}
A.~Rasouli, I.~Kotseruba, T.~Kunic, and J.~K. Tsotsos, ``{PIE}: A large-scale
  dataset and models for pedestrian intention estimation and trajectory
  prediction,'' in \emph{2019 IEEE/CVF International Conference on Computer
  Vision (ICCV)}, 2019, pp. 6261--6270.

\bibitem{8626436}
X.~Du, R.~Vasudevan, and M.~Johnson-Roberson, ``{Bio-LSTM}: A biomechanically
  inspired recurrent neural network for 3-d pedestrian pose and gait
  prediction,'' \emph{IEEE Robotics and Automation Letters}, vol.~4, no.~2, pp.
  1501--1508, 2019.

\bibitem{Liang_2019_CVPR}
J.~Liang, L.~Jiang, J.~C. Niebles, A.~G. Hauptmann, and L.~Fei-Fei, ``Peeking
  into the future: Predicting future person activities and locations in
  videos,'' in \emph{Proceedings of the IEEE/CVF Conference on Computer Vision
  and Pattern Recognition (CVPR)}, 2019.

\bibitem{Yuan2021AgentFormerAT}
Y.~Yuan, X.~Weng, Y.~Ou, and K.~M. Kitani, ``{AgentFormer}: Agent-aware
  transformers for socio-temporal multi-agent forecasting,'' in \emph{2021
  IEEE/CVF International Conference on Computer Vision (ICCV)}, 2021, pp.
  9793--9803.

\bibitem{7780479}
A.~Alahi, K.~Goel, V.~Ramanathan, A.~Robicquet, L.~Fei-Fei, and S.~Savarese,
  ``Social {LSTM}: Human trajectory prediction in crowded spaces,'' in
  \emph{2016 IEEE Conference on Computer Vision and Pattern Recognition
  (CVPR)}, 2016, pp. 961--971.

\bibitem{Vemula2017SocialAM}
A.~Vemula, K.~Muelling, and J.~Oh, ``Social attention: Modeling attention in
  human crowds,'' in \emph{2018 IEEE International Conference on Robotics and
  Automation (ICRA)}, 2018, pp. 4601--4607.

\bibitem{trajectron}
\BIBentryALTinterwordspacing
B.~Ivanovic and M.~Pavone, ``The trajectron: Probabilistic multi-agent
  trajectory modeling with dynamic spatiotemporal graphs,'' \emph{CoRR}, vol.
  abs/1810.05993, 2018. [Online]. Available:
  \url{https://arxiv.org/abs/1810.05993}
\BIBentrySTDinterwordspacing

\bibitem{9774877}
P.~R.~G. Cadena, Y.~Qian, C.~Wang, and M.~Yang, ``Pedestrian graph+: A fast
  pedestrian crossing prediction model based on graph convolutional networks,''
  \emph{IEEE Transactions on Intelligent Transportation Systems}, vol.~23,
  no.~11, pp. 21\,050--21\,061, 2022.

\bibitem{velickovic2018graph}
P.~Veli{\v{c}}kovi{\'c}, G.~Cucurull, A.~Casanova, A.~Romero, P.~Li{\`o}, and
  Y.~Bengio, ``Graph attention networks,'' in \emph{International Conference on
  Learning Representations (ICLR)}, 2018.

\bibitem{DBLP:journals/corr/abs-1803-10892}
A.~Gupta, J.~Johnson, L.~Fei-Fei, S.~Savarese, and A.~Alahi, ``Social {GAN}:
  Socially acceptable trajectories with generative adversarial networks,'' in
  \emph{Proceedings of the IEEE/CVF Conference on Computer Vision and Pattern
  Recognition (CVPR)}, 2018, pp. 2255--2264.

\bibitem{GAIL}
J.~Ho and S.~Ermon, ``Generative adversarial imitation learning,'' in
  \emph{Advances in Neural Information Processing Systems}, vol.~29, 2016.

\bibitem{sa_gail}
\BIBentryALTinterwordspacing
H.~Zou, H.~Su, S.~Song, and J.~Zhu, ``Understanding human behaviors in crowds
  by imitating the decision-making process,'' \emph{CoRR}, vol. abs/1801.08391,
  2018. [Online]. Available: \url{https://arxiv.org/abs/1801.08391}
\BIBentrySTDinterwordspacing

\bibitem{reinf}
R.~S. Sutton, D.~McAllester, S.~Singh, and Y.~Mansour, ``Policy gradient
  methods for reinforcement learning with function approximation,'' in
  \emph{Advances in Neural Information Processing Systems}, vol.~12, 1999, pp.
  1057--1063.

\bibitem{ucy}
A.~Lerner, Y.~Chrysanthou, and D.~Lischinski, ``Crowds by example,''
  \emph{Computer Graphics Forum}, vol.~26, no.~3, pp. 655--664, 2007.

\bibitem{PPO}
\BIBentryALTinterwordspacing
J.~Schulman, F.~Wolski, P.~Dhariwal, A.~Radford, and O.~Klimov, ``Proximal
  policy optimization algorithms,'' \emph{CoRR}, vol. abs/1707.06347, 2017.
  [Online]. Available: \url{https://arxiv.org/abs/1707.06347}
\BIBentrySTDinterwordspacing

\end{thebibliography}

\section{Appendices}
\subsection{Evaluation Metrics}
 In the context of the formulations of this paper, the ADE is an average squared geometric distance between the forecasted and the ground truth trajectories, which can be written as
\begin{align}
\text{ADE} = \mathbb{E}_{s_0 \sim b_0\left(s\right)} \Bigg[ \frac{1}{T_f^\tau} \mkern-4mu\sum_{t^\tau=T_o^\tau +1}^{T_o^\tau+ T_f^\tau}\mkern-13mu\big((\hat{x}_{t^\tau} - x_{t^\tau})^2 + (\hat{y}_{t^\tau} - y_{t^\tau})^2 \big) \Bigg].
\end{align}
The FDE considers only the final step of each pedestrian and it simplifies to
\begin{align}
\text{FDE} = \mathbb{E}_{s_0 \sim b_0\left(s\right)} \Big[(\hat{x}_{T_o^\tau+ T_f^\tau} - x_{T_o^\tau+ T_f^\tau})^2 \notag \\
+ (\hat{y}_{T_o^\tau+ T_f^\tau} - y_{T_o^\tau+ T_f^\tau})^2  \Big].
\end{align}
The coordinates are indexed by the trajectory time $t^\tau$.
\subsection{Hyperparameters Choice for Each Formulation}
The following table shows the chosen parameters, where the policy learning rate and the value function learning rate were the only parameters that had a noticeable impact on the performance and were used in the hyperparameter space searching. The PPO clipping factor was fixed at $0.2$. The value function iterations for PPO were kept at $1$. The learning rate is denoted as LR.

\begin{table}[h]
\vspace{-10pt}
    \centering
    \caption{Standard Learning Formulations}
    \label{tab:standard_learning}
    \begin{tabular}{@{}lc@{}}
        \toprule
        Formulation & Policy Learning Rate \\
        \midrule
        SL-MSE & $7.58 \times 10^{-5}$ \\
        SL-MSE-SPG & $8.62 \times 10^{-5}$ \\
        SL-MSE-SDM & $1.04 \times 10^{-4}$ \\
        \bottomrule
    \end{tabular}
\end{table}

\begin{table}[h]
\vspace{-20pt}
\setlength{\tabcolsep}{2pt}
    \centering
    \caption{Algorithms with value function}
    \label{tab:baseline_models}
    \begin{tabular}{@{}lcc@{}}
        \toprule
        Formulation & Policy LR & Value Function LR \\
        \midrule
        REINFORCE with Simplified Baseline & $9.5 \times 10^{-5}$ & $3.0 \times 10^{-4}$ \\
        REINFORCE with Full State Baseline & $2.32 \times 10^{-4}$ & $3.0 \times 10^{-4}$ \\
        \bottomrule
    \end{tabular}
\vspace{-20pt}
\end{table}


\begin{table}[H]
    \centering
    \caption{PPO-based algorithms}
    \setlength{\tabcolsep}{3pt}
    \label{tab:ppo_models}
    \begin{tabular}{@{}lccc@{}}
        \toprule
        Formulation & Policy LR & Value Function LR & PPO Iterations \\
        \midrule
        PPO - Simplified Baseline & $2.9 \times 10^{-5}$ & $1.0 \times 10^{-3}$ & 7 \\
        PPO - Full State Baseline & $2.8 \times 10^{-5}$ & $1.0 \times 10^{-3}$ & 7 \\
        PPO without Baseline & $7.8 \times 10^{-5}$ & - & 6 \\
        \bottomrule
    \end{tabular}
\end{table}


\subsection{SL-MSE-SDM Objective Function Derivation}
Following the definition of the objective function, for SL-MSE-SDM, the time horizon $T^\epsilon=12$ and the discount factor is now nonzero. The objective function can be calculated as
\begin{align}
\nabla_{\theta} J(\theta) & =\mathbb{E}_{s_0 \sim b_0\left(s\right)} \nabla_{\theta} V\left(s_0 ; \theta\right) = \notag \\
&=\mathbb{E}_{s_0 \sim b_0\left(s\right)}\mathbb{E}_{\left(\varepsilon \mid s_t=s_0\right) \sim \pi_\theta} \left[\nabla_{\theta}  G\left(\varepsilon \mid s_{t^\epsilon}=s_0\right)\right] \notag\\
& =\mathbb{E}_{s_0 \sim b_0\left(s\right),\left(\varepsilon \mid s_0\right) \sim \pi_\theta}\left[\nabla_{\theta} \sum_{t^\epsilon=0}^{T^{\varepsilon}-1} \gamma^{t^\epsilon} r_{t^\epsilon}\right] \notag\\
& =\mathbb{E}_{s_0 \sim b_0\left(s\right),\left(\varepsilon \mid s_0\right) \sim \pi_\theta}\mkern-6mu \left[ \nabla_{\theta}   \mkern-6mu \sum_{t^\epsilon=0}^{T^{\varepsilon}-1} \mkern-10mu\gamma^{t^\epsilon} \mkern-8mu\left\|\left(\hat{s}_{t^\epsilon}, \hat{a}_{t^\epsilon}\right)\mkern-4mu-\mkern-4mu\left(s_{t^\epsilon}, a_{t^\epsilon}\right)\right\|_2^2\right] \notag \\
&=\mathbb{E}_{s_0 \sim b_0\left(s\right),\left(\varepsilon \mid s_0\right) \sim \pi_\theta}\Bigg[ \nabla_{\theta}\sum_{t^\tau=T_o^\tau +1}^{T_o^\tau+ T_f^\tau}  \gamma^{k-T_o^\tau -1}  \notag
\sum_{l'=T_o^\tau+1 }^{t^\tau} \\
 & \left((\hat{x_{l'}} - x_{l'})^2 + (\hat{y_{l'}} - y_{l'})^2\right) \Bigg].
\end{align}

\subsection{SL-MSE-SPG Loss Function Derivation}
The policy gradient theorem \cite{reinf} uses so called parametrization and score function tricks to achieve the expression that uses logarithmic probabilities. Following the policy gradient theorem, $T^\epsilon=1$ and $\lambda=0$ the derivation follows as 
\begin{align}
\nabla_{\theta}J(\theta) & =\mathbb{E}_{s_0 \sim b_0\left(s\right)} \nabla_{\theta}V\left(s_0 ; \theta\right) \notag \\
& =\mathbb{E}_{s_0 \sim b_0\left(s\right),\left(\varepsilon \mid s_0\right) \sim \pi_\theta}\left[\nabla_{\theta} \sum_{t^\epsilon=0}^{T^{\varepsilon}-1} \gamma^{t^\epsilon} r_t\right] \notag\\
& =\mathbb{E}_{s_0 \sim b_0\left(s\right),\left(\varepsilon \mid s_0\right) \sim \pi_\theta}\mkern-6mu \left[ \nabla_{\theta}   \mkern-6mu \sum_{t^\epsilon=0}^{T^{\varepsilon}-1} \mkern-10mu\gamma^{t^\epsilon} \mkern-8mu\left\|\left(\hat{s}_{t^\epsilon}, \hat{a}_{t^\epsilon}\right)\mkern-4mu-\mkern-4mu\left(s_{t^\epsilon}, a_{t^\epsilon}\right)\right\|_2^2\right] \notag \\
& =\mathbb{E}_{s_0 \sim b_0(s),\, (\varepsilon \mid s_0) \sim \pi_\theta}\Bigg[ \sum_{t^\tau=T_o^\tau +1}^{T_o^\tau+ T_f^\tau} \mkern-13mu\Big((\hat{x}_{t^\tau} - x_{t^\tau})^2 + \notag \\
& (\hat{y}_{t^\tau} - y_{t^\tau})^2 \Big) \times \nabla_{\theta} \left( \sum_{t^\tau=T_o^\tau +1}^{T_o^\tau+ T_f^\tau} \log \pi_{\theta} \left( \hat{a_{t^\tau}} \mid \hat{s_{t^\tau}} \right) \right) \Bigg].
\end{align}
\end{document}